\documentclass[3p,times,procedia]{elsarticle}
\usepackage{csquotes}

\usepackage{ecrc}
\usepackage[bookmarks=false]{hyperref}
    \hypersetup{colorlinks,
      linkcolor=blue,
      citecolor=blue,
      urlcolor=blue}

\volume{00}

\firstpage{1}

\journalname{Procedia Computer Science}

\runauth{C. Creangă et.al.}

\jid{procs}

\usepackage{amssymb}

\usepackage[figuresright]{rotating}

\begin{document}
\begin{frontmatter}



\dochead{28th International Conference on Knowledge-Based and Intelligent Information \& Engineering Systems (KES 2024)}%

\title{Fine-Tuning Models for Biomedical Relation Extraction}

\author[a,c]{Claudiu Creangă} 
\author[b,c]{Liviu P. Dinu}
\author[d]{Daniela Gifu}

\address[a]{Interdisciplinary School of Doctoral Studies}
\address[b]{Faculty of Mathematics and Computer Science, University of Bucharest, Romania}
\address[c]{HLT Research Center}
\address[d]{Institute of Computer Science, Romanian Academy, Iasi Branch}

\begin{abstract}
Next-Generation Sequencing has revolutionized the study of genetic mutations, enabling large-scale investigations into their roles in disease development. However, extracting meaningful insights from the vast amount of biomedical literature remains a complex challenge that cannot be addressed manually. In this paper, we present pre-trained models (PTMs) for the automatic extraction of relations from biomedical text, specifically targeting the variant-phenotype domain. Our evaluation on the SNPPhenA corpus demonstrates that fine-tuning small BERT-based models, particularly DeBERTa, yields strong performance, approaching the current state-of-the-art (SOTA). Additionally, our results indicate that carefully fine-tuning Google’s Gemini Pro 1.0 outperforms the existing SOTA for both sentence-level tasks (where the model processes only the target sentence) and abstract-level tasks (where the model processes the entire abstract).
\end{abstract}


\end{frontmatter}

\email{claudiu.creanga@s.unibuc.ro}



\section{Introduction}
\label{introduction}

Next-Generation Sequencing (NGS) has revolutionized biomedical research, playing a pivotal role in understanding genetic mutations \cite{Satam2023}, which are integral of the propagation of the human species \cite{Gifu2019}. This advanced sequencing technology has enabled the way scientists explore DNA and RNA, offering an unprecedented scale of analysis. By enabling high-throughput sequencing, NGS generates vast amounts of data for understanding the complex landscape of genetic mutations. Researchers harness NGS to investigate the genomes of individuals affected by genetic disorders, cancers, and inherited conditions. Whether sequencing entire genomes or targeting specific regions, NGS facilitates the identification of various genetic alterations. These include single-nucleotide variants (SNPs or SNVs), indels (deletions or insertions), copy number alterations or variations (CNVs), and structural rearrangements (SVs), shedding light on their roles in disease development and progression. Note that sometimes SNVs are known as single nucleotide polymorphisms (SNPs), although SNV and SNPs are not interchangeable. NGS has facilitated large-scale population studies, where the genomes of thousands or even millions of individuals are sequenced. By comparing the genetic makeup of healthy individuals with those affected by specific diseases, researchers can identify genetic variations that confer susceptibility to certain conditions.

The question of this study is: \textit{How powerful are NLP tools in using the NGS data from genomics research?} 

Given the huge number of scientific articles published in biomedical journals, triggered by the NGS revolution, it is impossible to manually curate the articles and build curated databases that are up to date \cite{Badal2019}. To solve this problem, NLP tools can automatically parse and extract meaningful data from biomedical text.  In this research, we introduce a deep learning approach utilizing transformers capable of extracting relations between entities from biomedical text. Our tool is evaluated on the SNPPhenA corpus \cite{Bokharaeian2017}. The null hypothesis of our research is that fine-tuned large language models don't improve the results on REL tasks, specifically on the SNPPhenA corpus. We will test this hypothesis and show that fine-tuned LLMs perform over the current state of the art. The rest of the paper is organized as follows: section 2 briefly presents studies related to the relation extraction (REL) process, section 3 provides information about the system architecture and training methods, section 4 resumes the results of the conducted experiments based on LLMs, with their interpretations, followed by section 5 with the conclusions.

\section{Related Work}
\label{related}
The success of NGS technology lies in its ability to sequence millions of DNA reads on a large scale, enabling multi-gene analysis with minimal nucleic acid input. Regarding the process of relation extraction (REL) from biomedical text, it typically involves three steps:

\begin{itemize}[]
\item \textbf{Named Entity Recognition (NER):} NER techniques are used to identify and classify entities such as genes, mutations, proteins, diseases, and drugs. These recognized entities serve as key participants in the subsequent relation classification task. A proficient NER tool is essential for accurately identifying the entities. 
\item \textbf{Relation Extraction (REL):} Once the entities are identified, relation extraction techniques are applied to discern the sentence’s structure and identify the relationships between entities. This process may entail analyzing grammatical patterns and dependency relationships within the entire sentence to accurately extract the relation.
\item \textbf{Relation Classification:} The final step involves classifying the extracted relations into predefined categories. These categories typically relate to the domain of the relation, such as determining if it represents a gene-disease relation or a drug-drug relation. 
\end{itemize}
 
Most of NER tools were rule-based, but recently, machine learning have significantly outperformed them. Rule-based models utilize POS tagging and parse trees, from which researchers construct manual rules to identify relations between entities \cite{Gifu2014}. These rules usually look for a dictionary of verbs like “treats” or “cures”. One of the most maintained tools is tmVar, which is now at version 3 \cite{Wei2022}. It uses a probabilistic model to extract entities. This tool is used by PubTator and LitVar on the official NCBI server \cite{LitVar}. Other classical tools are PKDE4J (rule-based, supervised) \cite{Verspoor2016} and DiMex  (rule-based, unsupervised) \cite{Ashique2019}. The current direction in research is to use pre-trained models based on transformers which achieve better results, like SciBERT \cite{Beltagy2019} and PubMedBERT \cite{Tinn2020}. 

The BLURB (Biomedical Language Understanding and Reasoning Benchmark) is a leader board in BioNLP developed by Microsoft. It serves as a platform that showcases the performance and progress of various natural language processing (NLP) models in the realm of biomedical text mining. This leader board acts as a benchmarking system, enabling researchers and developers to compare the efficacy of different NLP techniques for addressing bioinformatics and biomedical research challenges. BLURB encompasses thirteen publicly available datasets across six diverse tasks \cite{Microsoft}. The six tasks are NER, PICO, REL, Sentence Similarity, Document Classification and Question Answering. 

\section{System Overview}

The tools presented in this survey enable the creation of large genetic mutation-phenotype databases without manual intervention, that are not only useful to use, but can also lead to new insights.

\subsection{Dataset}
In this research, we focus on Relation Extraction, and BLURB employs three datasets: ChemProt, which contains chemical-protein annotations and protein-protein interactions; DDI, which contains drug-disease interactions; and GAD, which contains gene-disease associations. For the REL task, the current SOTA scores are displayed in \autoref{table:REL}.

\begin{table}[h]
\caption{BLURB leaderboard for REL task as of April 2024. The F1 score is a measure of a model's performance. It is a way of combining the precision and recall of the model, and it is defined as 2 * (precision * recall) / (precision + recall).}
\begin{tabular*}{\hsize}{@{\extracolsep{\fill}}lll@{}}
\toprule
Model & F1 score \\
\colrule
BioM-BERT-PMC-Large & 83.27 \\
BioM-ELECTRA-Large  &  83.17\\
BioM-ALBERT-xxlarge-PMC  &  83.14\\
\botrule
\label{table:REL}
\end{tabular*}
\end{table}


In case of the REL task, historically, models were based on co-occurrence: entities serve as vertices in a graph, and the relations between them represent the edges. Note that if two entities appear in the same sentence, a connection between them is logged. If these entities frequently appear together, the connection is deemed stronger. The advantages of this approach include:
\begin{itemize}[]
\item \textbf{Simplicity:} Co-occurrence models are relatively easy to implement, understand, and interpret compared to more complex models, such as deep learning models.
\item \textbf{Efficiency:} These models can be highly efficient, particularly for large corpora, because they don't require intensive computational resources and can be parallelized easily.
\item \textbf{No need for annotated data:} Co-occurrence models don't require labelled data, making them a good choice when labelled data is scarce. They work based on statistical properties of the data.
\end{itemize}

There are obvious limitations to this approach, as two entities may appear in the same sentence, but the relation between them can be negated by a clause. Co-occurrence models lack an understanding of the underlying semantics of the text; they only recognize that two entities frequently appear together, without comprehending how or why they are related. Consequently, these models can be susceptible to noise, inferring a strong relationship between entities solely based on their frequent co-occurrence, even if the relationship is coincidental or misleading. Additionally, co-occurrence models struggle to capture complex, non-linear, or long-distance relationships between entities. They typically only capture relationships within a fixed window of context, which is particularly problematic in biomedical papers where relationships can span multiple sentences. Moreover, co-occurrence models heavily rely on term frequency, which may not always align with the semantic importance of terms in biomedicine. In biomedical papers, redundant information is typically absent, meaning important statements are not repeated.

\textbf{Rule-based approaches} start with POS tagging, construct parse trees and then manually figure out rules (with the help of domain experts) to find the relationship. The advantages of the rule-based methods are: 
\begin{itemize}[]
\item \textbf{Interpretability:} You know exactly why the model is making a certain prediction because it's following a specific rule that you've defined. If something goes wrong, you know where and what to change.
\item \textbf{No need for annotated data:} Rule-based models do not require annotated training data, which can be costly or time-consuming to gather. You only need to define the rules.
\item \textbf{High precision:} If the rules are carefully and correctly defined, rule-based models can achieve high precision, because they only predict a relationship when a specific pattern is matched.
\end{itemize}

However, rule-based methods are not as prevalent nowadays due to the significant manual effort required to build them. Experts who annotate biomedical papers are expensive and scarce resources. Creating these rules demands a considerable amount of time and effort. Additionally, rule-based models may suffer from low recall, meaning they might miss many relationships that do not conform to the pre-defined patterns. Rules that are effective in one text or domain may not be suitable for another. Furthermore, these models may struggle to generalize well to new texts or domains without additional rule-making efforts. Ultimately, even for experts, extracting generalizable rules from ambiguous natural language is challenging. One notable project that utilizes rule-based algorithms is PKDE4J\cite{Verspoor2016}.

Today, the most performant models are based on deep learning (CNN, LSTM and BERT). Deep learning models have the capacity to automatically learn complex patterns, dependencies, and relationships within data. They excel at capturing non-linear and long-distance dependencies between words. These models can autonomously extract features and predict relations in an end-to-end manner, eliminating the need for manually crafted features or rules. One major advantage of deep learning is its ability to leverage transfer learning. Pre-trained models such as BERT, trained on large datasets sourced from the web or other domains, can be fine-tuned on specific tasks like relation extraction using smaller, task-specific datasets. This approach has yielded state-of-the-art results across numerous tasks. However, deep learning also comes with its own challenges, including the requirement for large amounts of annotated data, significant computational resources, and the risk of overfitting if training is not properly regularized or if the dataset is too small or imbalanced. Within the realm of relation extraction (REL), there are several sub-domains depending on the subject matter of the text, such as drug-disease, gene-disease, chemical-protein, and variant-phenotype relations. In this research, we will concentrate on the variant-phenotype relation, which has received less attention compared to others. Specifically, there is only one dataset available that annotates associations between Single Nucleotide Polymorphisms (SNPs) and phenotypes, known as the SNPPhenA corpus \cite{Bokharaeian2017}. The dataset has received little attention, only three published papers used it, as far as we are aware (\cite{Bokharaeian2017}, \cite{Dehghani2023}, \cite{Bokharaeian2023}). The corpus was manually curated by three experts and is regarded as being done to a high standard. To better understand the variant-phenotype relation extraction task we supply an example of an abstract from a biomedical text: 

\begin{displayquote}
LD (r2=0.35) between TOMM40 (rs2075650) and APOC1 (rs1064725) was observed in PPA, but not in controls and in bvFTD. Inside this region of 26.9 kb, LD (r2 \textgreater 0.50) between TOMM40 (rs2075650) and APOE (rs429358) was observed in bvFTD and in controls, but not in PPA. Inside this region of 16.3 kb, LD (r2 = 0.14) between TOMM40 (rs157590) and APOE (rs429358) was observed in PPA, but not in bvFTD and in controls (emphasis mine) \cite{Seripa2012}.
\end{displayquote}

\begin{figure}[t]\vspace*{3pt}
\centerline{\includegraphics[width=0.8\linewidth]{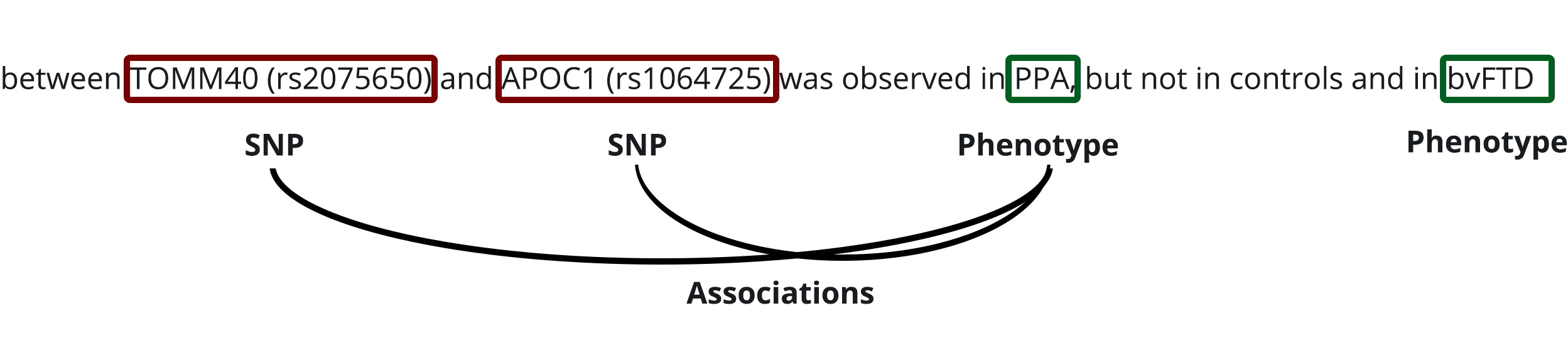}}
\caption{ Association between 2 SNPs and one phenotype, ignoring the other phenotype. \cite{Bokharaeian2017} }
\label{fig:association}
\end{figure}

We can observe three sentences that mention a relationship between one or two SNPs and a phenotype. In that abstract, the SNPs mentioned are rs1064725, rs429358, rs157590, and rs429358, and the phenotypes mentioned are PPA (Primary progressive aphasia) and bvFTD (Behavioral variant frontotemporal dementia). These entities are extracted using an NER tool. Based on the provided abstract, our REL model builds a system that detects the relation as in \autoref{fig:association}.

The current SOTA results for the SNPPhenA corpus at sentence-level are shown in \autoref{table:SNPPhrenA}. 

\begin{table}[h]
\caption{SOTA for SNPPhenA corpus (sentence level).}
\begin{tabular*}{\hsize}{@{\extracolsep{\fill}}llll@{}}
\toprule
Model & Precision & Recall & F1 score (macro) \\
\colrule
CNN-LSTM & 0.732 & 0.723 & 0.723 \\
BERT-LSTM & 0.739 & 0.803 & 0.730\\
PubMedBERT-LSTM & 0.870 & 0.883 & 0.866\\
BioBERTGRU & \textbf{0.883} & \textbf{0.882} & \textbf{0.881} \\
\botrule
\label{table:SNPPhrenA}
\end{tabular*}
\end{table}

\subsection{BERT-based models}
\begin{figure}[t]\vspace*{3pt}
\centerline{\includegraphics[width=0.8\linewidth]{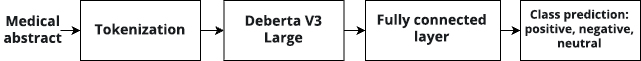}}
\caption{ Model architecture based on DeBERTa. }
\label{fig:model}
\end{figure}

MLMs learn by strategically masking, or hiding, words within a sentence and predicting the missing words based on the surrounding context (bi-directional). This differs from Causal models, which are autoregressive and predict the next token based solely on the previous tokens. We experimented with several MLMs (as observed in \autoref{table:deberta}), DeBERTa \cite{he2021debertav3} being the best model, according to this architecture. On top of this pre-trained model we use a fully connected layer as the output layer composed of: normalization layer, tanh activation layer, a dropout layer and a final linear layer with output size 1 for classification, as seen in \autoref{fig:model}. 

The strategy was not to use any preprocessing steps (stop word removal or lemmatization) (such as stop word removal or lemmatization), similar to what was done in the paper that achieved the current best results \cite{Dehghani2023}. 
Instead, our results suggest that BERT-based models perform better with natural language as is, and pre-processing it removes informative features, not just noise. The length of the samples in the corpus is low, so a maximum sequence length of 512 covers all the examples.

\subsection{Layer Selection}
These experiments revealed that fine-tuning the pre-trained model using only the last layer yielded the best results (see \autoref{table:layer}). This suggests that the final layer may contain the most relevant features for our specific task.

\begin{table}[h]
\caption{We use F1 score (macro) to optimize layer selection from the pre-trained model (sentence-level task). Here's our convention: -1 indicates the last layer, -2 the second-to-last, and {[}-1, -3{]} signifies the concatenation of the final three layers. Performance was evaluated on a validation set consisting of 20\% of the authors' original training data, yielding the following results. }
\begin{tabular*}{\hsize}{@{\extracolsep{\fill}}llll@{}}
\toprule
Layer & F1 score (macro) \\
\colrule
-1 & 0.85  \\
-2 & 0.80\\
-3 & 0.78 \\
{[}-1, -3{]} & 0.83 \\
\botrule
\label{table:layer}
\end{tabular*}
\end{table}

\subsection{Fine-Tuning}

Each layer within the transformer architecture processes the input text, extracting increasingly complex patterns. Lower layers focus on basic syntactic structures, while higher layers capture more nuanced semantic relationships between words and phrases \cite{Yosinski2014}. To fine-tune the layers, we followed the two-step approach presented in \cite{Sun2019}. 

First, for a set number of epochs \textit{k} we froze the pre-trained layers and only trained the fully connected layer, which sits on top of the network. In this step, we utilized a higher learning rate of 5e-5 and a warm-up period of 10,000 steps with the AdamW optimizer.

Second, for a number of epochs \textit{j} we train the selected layer from the pre-trained model (the last layer) and the fully connected layer. Therefore, we update the weights of the pre-trained model as well. Because this step can easily lead to overfitting, especially given the small size of the SNPPhenA corpus, we use a lower learning rate (2e-5). By following this two-step approach and freezing a large portion of the pre-trained model, we achieve several benefits: conserving computational resources, preventing overfitting, and enabling the model to retain more generalized task knowledge. We tested the fine-tuning strategy together with the layer selection. As we selected more layers to fine-tune, the performance degraded as the model started to overfit on the dominant class. When fine-tuning the whole model, without freezing any layers, we obtained a F1-score of 0.66.  

\subsection{Causal Models}

Besides MLMs, we tested the following open source Causal Models: SOLAR-10.7B-Instruct-v1.0 \cite{kim2023solar}, Mistral 7 \cite{jiang2023mistral}, LLama-2 7B \cite{touvron2023llama} and the proprietary Gemini 1.0 Pro and Gemini 1.5 Pro from Google. These models were used in a zero-shot setting and in a few-shot setting. In the zero-shot setting we provided the model with the sample to test and ask it to predict the class, without any training data or examples. The following prompt was the one that gave us the best results in case of open source models: 
\begin{displayquote}
"[INS] Given the following sentence: \verb|{sentence}|. Predict the relation between the SNP \verb|{snp} | and the phenotype \verb|{phenotype}|. Chose one of the following options: positive, neutral, negative. Answer in one word only.  Answer: [\verb|\INS|]"
\end{displayquote}

In the few-shot setting we gave the model a few examples (one from each class) in the same prompt on how to solve the task. This is the prompt that worked best:

\begin{displayquote}
    Q: Given the following sentence: "RESULTS: Over an average period of three years, participants with the risk-conferring TT genotype at rs7903146 were more likely to have progression from impaired glucose tolerance to diabetes than were CC homozygotes (hazard ratio, 1.55; 95 percent confidence interval, 1.20 to 2.01; P<0.001)." Predict the association between the SNP "rs7903146" and the phenotype "diabetes". Chose one of the following options: positive, neutral, negative. Chose positive only if there is a clear association. Chose negative only if the association is negated. Otherwise chose neutral. Answer in one word only. Think about it step by step.
    
    A: positive
    
    Q: Given the following sentence: "Although there was only moderate linkage disequilibrium between rs2075650 and the ApoE E4 defining SNP rs429358, we could not find an APOE-independent effect of rs2075650 on longevity, either in cross-sectional or in longitudinal analyses." Predict the association between the SNP "rs2075650" and the phenotype "longevity". Chose one of the following options: positive, neutral, negative. Chose positive only if there is a clear association. Chose negative only if the association is negated. Otherwise chose neutral. Answer in one word only. Think about it step by step.
    
    A: negative
    
    Q: Given the following sentence: "CONCLUSIONS: The frequency of ApoE polymorphisms did not differ between patients with RA and controls." Predict the association between the SNP "ApoE polymorphisms" and the phenotype "RA". Chose positive only if there is a clear association. Chose negative only if the association is negated. Otherwise chose neutral. Answer in one word only. Think about it step by step.
    
    A: neutral
        
\end{displayquote}

In our experiment, we aimed to guide a large language model (LLM) to break down complex problems into smaller, more manageable steps. We achieved this by prompting the LLM with the phrase "think about it step by step." However, simply instructing the LLM to break down problems wasn't enough. We found that providing additional examples and prompts that encouraged the LLM to explicitly show its reasoning process (chain of thought prompting) were crucial for optimal performance. Interestingly, adding another step, "explain your decision," which is a common recommendation in prompt engineering, did not improve performance and even led to a decrease (as shown in \autoref{table:causal}). This suggests that while explaining reasoning can be helpful in some situations, it might not be universally beneficial for LLMs.

Google allows users to fine-tune the model Gemini 1.0 Pro. Two fine-tuning experiments were conducted: (1) the complete training dataset and (2) a subset comprised of 50 randomly selected elements from each class. The fine-tuning process involved 2 epochs and a learning rate multiplier of 1.

\section{Results}

We could test our models at sentence level or abstract level. In case of MLMs and open source Casual Models we chose to test it only at sentence level because it required less resources. In case of the fine-tuned Gemini 1.0 Pro, we tested it also at abstract level. 

\subsection{Performance of MLMs at sentence level}

We obtained good results using fine-tuned MLMs, see \autoref{table:deberta}. We tested 4 pre-trained models using our architecture and fine-tuning process. The model that performed best was based on DeBERTa, which obtained an F1 score of 0.85, close to the state of the art model of BioBERTGRU (0.88). The model is smaller than  BioBERTGRU, but still achieved great accuracy. 

\begin{table}[h]
\caption{Performance of MLMs based on our architecture (test corpus - sentence level).}
\begin{tabular*}{\hsize}{@{\extracolsep{\fill}}llll@{}}
\toprule
Model & F1 score (macro) \\
\colrule
DeBERTa v3 & 0.85  \\
BioBERT & 0.80\\
BERT & 0.73 \\
DistilBERT & 0.67 \\
\botrule
\label{table:deberta}
\end{tabular*}
\end{table}

\subsection{Performance of Causal Models at sentence level}

As we can see in \autoref{table:causal}, the best open source Causal model is SOLAR-10.7B-Instruct-v1.0. Unfortunately it achieved a lower F1 score (0.67) than the standard CNN-LSTM model (see \autoref{table:SNPPhrenA}). In general, for smaller Casual Models, the zero-shot strategy performed better than the few-shot one. This suggests that the model struggled with longer text and forgot important information. 

\begin{table}[h]
\caption{Performance of Causal Models on the test corpus at sentence level. The environment mentions if it was tested in a zero-shot setting or a few-shot setting or if it was fine-tuned using the training dataset. CoT1 means that the chain of thought phrase "think about it step by step" was used or not. CoT2 means that the chain of thought phrase "explain your decision" was used or not in addition to the "think about it step by step" phrase. Fine-tuning was done either with the full training dataset or with a balanced training dataset of 50 examples from each class. Then we tested the fine-tuned model on the whole test corpus. }
\begin{tabular*}{\hsize}{@{\extracolsep{\fill}}llll@{}}
\toprule
Model & Environment & F1 score (macro) \\
\colrule
Gemini 1.0 Pro & Fine-tuned - balanced train dataset, CoT1 & 0.89 \\
Gemini 1.5 Pro & Few-shot, CoT1 & 0.77 \\
Gemini 1.0 Pro & Fine-tuned - full train dataset, CoT1 & 0.72  \\
Gemini 1.0 Pro & Fine-tuned - full train dataset, CoT2 & 0.70 \\
Gemini 1.0 Pro & Few-shot, CoT1 & 0.70 \\
Solar 10B & Zero-shot & 0.67 \\
Mistral 7B & Zero-shot & 0.66  \\
Solar 10B & Few-shot & 0.66 \\
Mistral 7B & Few-shot & 0.65  \\
Gemini 1.0 Pro & Zero-shot, CoT1 & 0.65 \\
Gemini 1.0 Pro & Zero-shot & 0.63 \\
Llama-2 7B & Zero-shot, CoT1 & 0.61\\
Llama-2 7B & Few-shot, CoT1 & 0.59\\
\botrule
\label{table:causal}
\end{tabular*}
\end{table}

Gemini 1.0 Pro, which is much larger than Solar 10B, did only slightly better, with a score of 0.7. A bigger difference was seen with the latest model from Google, Gemini 1.5 Pro which achieved an F1 score of 0.77 in a few-shot setting. Google doesn't allow to fine-tune Gemini 1.5 Pro, but we can fine-tune the Gemini 1.0 Pro model and it achieved new state of the art scores on the SNPPhenA corpus. Our training dataset was imbalanced, containing a much higher proportion of positive examples compared to negative or neutral ones. To prevent overfitting on the majority class, we experimented with training the model on a smaller subset containing only 50 examples from each class. Surprisingly, this approach yielded the best results, achieving a score of 0.89. This outperforms previous models like BioBERT-GRU and PubMedBERT-LSTM (previous SOTA). Additionally, training on a smaller dataset suggests the model might generalize better to unseen data.

\begin{figure}[t]\vspace*{4pt}
\centerline{\includegraphics[width=0.9\linewidth]{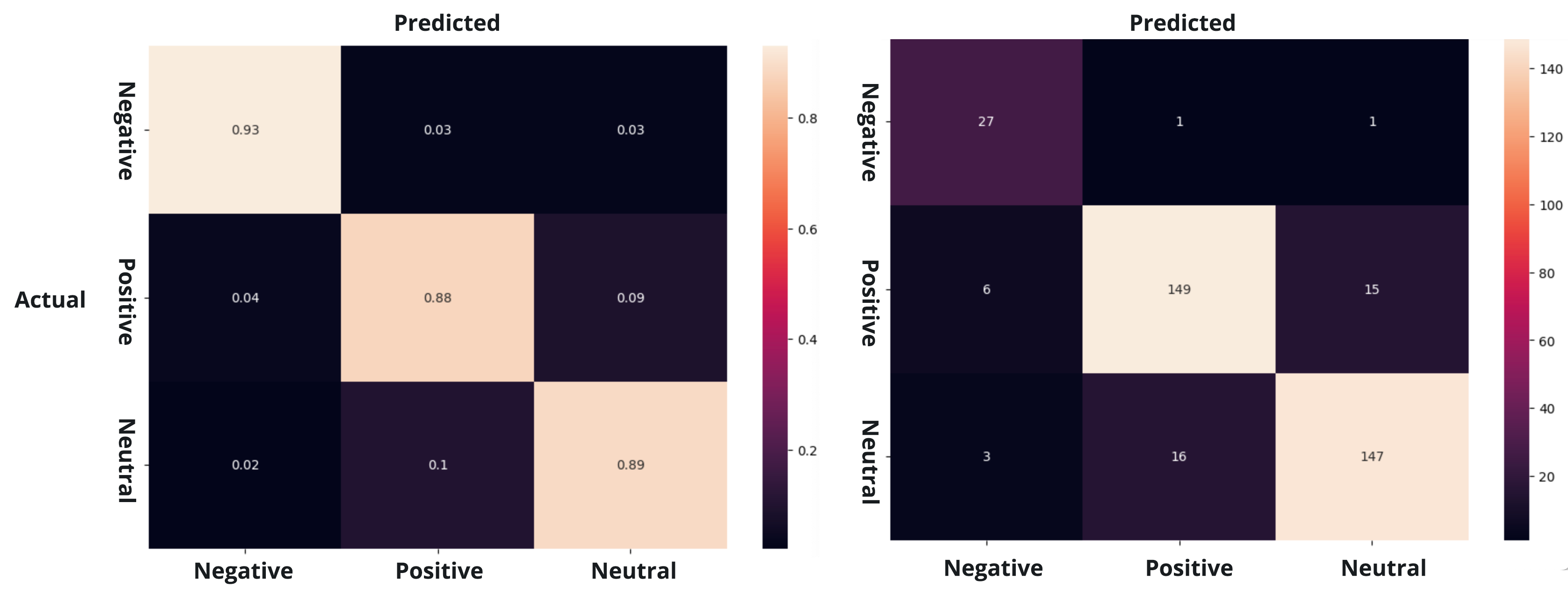}}
\caption{ Confusion matrix of our best model, the fine-tuned version based on Gemini 1.0 Pro. The values on the left are normalised and the values on the right are raw. Due to class imbalance in the dataset, we normalized the confusion matrix by dividing each value by its row or column total.}
\label{fig:confusion_matrix}
\end{figure}

From the confusion matrix (see \autoref{fig:confusion_matrix}), we can deduct that the model is performing very well, most predictions falling along the diagonal. It appears there's more confusion with the positive label, as it has relatively more misclassifications compared to the negative and neutral labels. As seen in \autoref{table:precision_recall}, the negative class has the lowest F1-score and a noticeable difference between precision and recall (we get the worst precision, 0.75). In exchange, we get very good precision for the neutral and positive class. 

\begin{table}[h]
\caption{Scores of the Gemini 1.0 Pro fine-tuned model per class (test corpus - sentence level). The test corpus is imbalanced and is intentionally designed with a different distribution than the training corpus, in order to identify models that can effectively generalize beyond the specific examples they've learned. }
\begin{tabular*}{\hsize}{@{\extracolsep{\fill}}llll@{}}
\toprule
Class & Precision & Recall & F1 score (macro) \\
\colrule
Positive (170 items) & 0.898 & 0.876 & 0.887 \\
Negative (29 items) & 0.750 & 0.931 & 0.831 \\
Neutral (166 items) & 0.902 & 0.886 & 0.894 \\
\botrule
\label{table:precision_recall}
\end{tabular*}
\end{table}

\subsection{Performance of Causal Models at abstract level}

\begin{table}[h]
\caption{Performance of Gemini models on the test corpus at abstract level. CoT means that the chain of thought phrase "think about it step by step" was used or not. Fine-tuning was done with a balanced training dataset of 50 examples from each class.}
\begin{tabular*}{\hsize}{@{\extracolsep{\fill}}llll@{}}
\toprule
Model & Environment & F1 score (macro) \\
\colrule
Gemini 1.0 Pro & Fine-tuned - balanced train dataset, CoT & 0.80 \\
Gemini 1.5 Pro & Few-shot, CoT & 0.73 \\
BioBERTGRU & previous SOTA & 0.645 \\
\botrule
\label{table:causal_abstract}
\end{tabular*}
\end{table}

\begin{figure}[t]\vspace*{3pt}
\centerline{\includegraphics[width=0.5\linewidth]{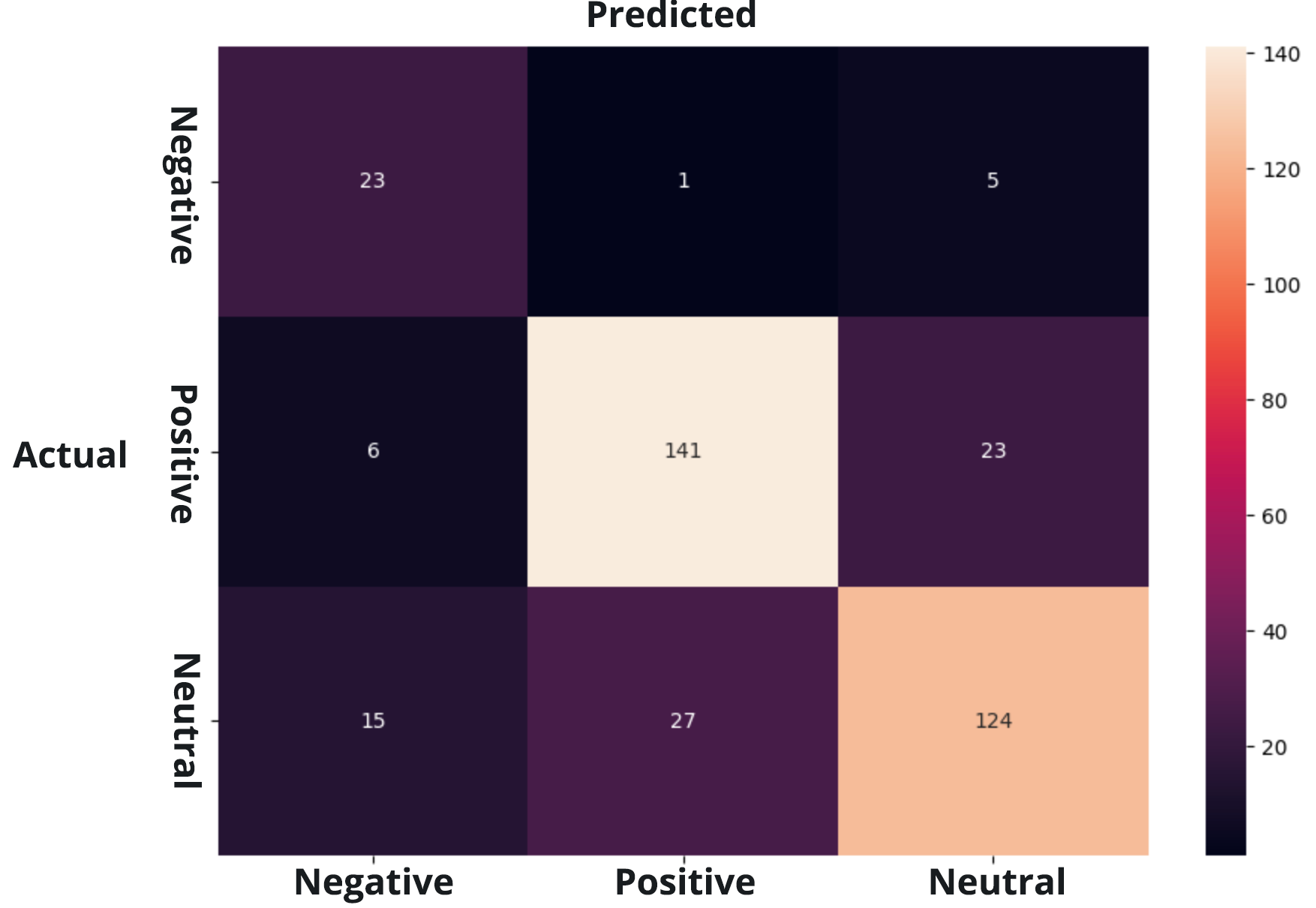}}
\caption{Confusion Matrix: Fine-tuned Gemini 1.0 Pro (Abstract-level Task). Values are calculated on the test dataset.}
\label{fig:confusion_matrix_abstract}
\end{figure}

We tested Google models at abstract level because these performed best at sentence level. With the fine-tune version of Gemini 1.0 Pro we obtained a new state of the art result by a large margin, 0.80 vs. 0.645 which was the previous best result (see \autoref{table:causal_abstract}). This suggests that, for this task, new LLMs perform much better than MLMs when the text is longer. Gemini 1.5 Pro, tested in a few-shot setting, also obtained a better score than the current SOTA, 0.73. Looking at the \autoref{table:precision_recall_abstract} we can see that again the negative class has the worst performance. Probably the model has issues with negation and the various ways in which negation can be expressed in language and which can flip the meaning of a sentence. The confusion matrix (see \autoref{fig:confusion_matrix_abstract}) reveals difficulty in distinguishing the neutral class, with 15 instances misclassified as negative, when they were neutral and 27 examples misclassified as positive, when they were neutral. 

\begin{table}[h]
\caption{Scores of the Gemini 1.0 Pro fine-tuned model per class (test corpus - abstract level). }
\begin{tabular*}{\hsize}{@{\extracolsep{\fill}}llll@{}}
\toprule
Class & Precision & Recall & F1 score (macro) \\
\colrule
Positive (170 items) & 0.834 & 0.829 & 0.832 \\
Negative (29 items) & 0.523 & 0.793 & 0.630 \\
Neutral (166 items) & 0.816 & 0.747 & 0.780 \\
\botrule
\label{table:precision_recall_abstract}
\end{tabular*}
\end{table}

\section{Conclusion}

This research highlights the significant role of NGS in genomics research, specifically its application for variant-phenotype relation extraction using both BERT-based models and Causal Models. Fine-tuning pre-trained MLMs, particularly DeBERTa, achieved strong performance on the SNPPhenA corpus, approaching the current state-of-the-art (SOTA). Our experiments demonstrated that strategically fine-tuning in two steps with different learning rates and freezing the weights except for the final layer of the pre-trained model yielded good results. While investigating, open-source Causal Language Models underperformed in this task, indicating they may struggle to grasp the nuances of variant-phenotype relationships in their zero-shot or few-shot setting. In contrast, proprietary Google models, Gemini Pro 1.0 and 1.5, performed better. The fine-tuning version of Gemini Pro 1.0 surpassed the SOTA in both sentence and abstract-level tasks. This suggests that large models, when fine-tuned, perform better than MLMs at detecting associations in the SNPPhenA corpus.

\section*{Acknowledgements}

This work was partially supported by a grant on Machine Reading Comprehension from Accenture Labs and by the POCIDIF project in Action 1.2. "Romanian Hub for Artificial Intelligence".








\clearpage

\end{document}